%% file: main_v3.tex
\documentclass[10pt,twocolumn,letterpaper]{article}

\usepackage[pagenumbers]{wacv}
\input{preamble}
\input{paper_info}

\definecolor{wacvblue}{rgb}{0.21,0.49,0.74}
\usepackage[breaklinks,colorlinks,allcolors=wacvblue]{hyperref}
\hypersetup{pdftitle={\paperTitle},pdfauthor={\paperAuthorNames}}

\title{\paperTitle}
\author{\paperAuthors}

\begin{document}
\maketitle
\raggedbottom
\widowpenalty=10000
\displaywidowpenalty=10000
\input{sec/v3_0_abstract}
\input{sec/v3_1_intro}
\input{sec/v3_2_related_work}
\input{sec/v3_3_operational_task}
\input{sec/v3_4_model}

\input{sec/v3_5_evaluation}
\input{sec/v3_6_results}
\input{sec/v3_7_discussion}
\input{sec/v3_8_conclusion}
\input{sec/v3_code_availability}
\input{sec/v3_9_acknowledgments}
{
    \small
    \bibliographystyle{ieeenat_fullname}
    \bibliography{main}
}
\end{document}


\maketitle
\raggedbottom
\widowpenalty=10000
\displaywidowpenalty=10000

\input{supplementary_sections/revised_protocols}
\input{supplementary_sections/revised_controls}

\input{supplementary_sections/homogeneous_failures}
\input{supplementary_sections/revised_qualitative}

%% file: preamble.tex
\usepackage{amsmath}
\usepackage{booktabs}
\usepackage{float}
\usepackage{microtype}
\usepackage{tabularx}

\newcommand{\TV}{\operatorname{TV}}

%% file: paper_info.tex
\newcommand{\paperTitle}{SteelHuman: Predicting Operator Estimates of Steel-Scrap Composition from Sparse RGB Views}

\newcommand{\paperAuthorNames}{Jovan Kamchev and Zoran Ivanovski}

\newcommand{\paperAuthors}{%
  Jovan Kamchev \qquad Zoran Ivanovski\\
  Faculty of Electrical Engineering and Information Technologies\\
  Ss. Cyril and Methodius University\\
  Skopje, North Macedonia\\
  {\tt\small jovankamchev@gmail.com}
}

%% file: sec/v3_0_abstract.tex
\begin{abstract}
Steel scrap composition estimation is an important part of yield planning 
and fair supplier compensation. Measuring the composition of a multi-ton pile is an inherently laborious process, so the assessment is often left to trained human operators. 
Recent works tackle composition-level tasks, but require structured acquisition and multi-stage preprocessing. 
We study a simpler approach to steel scrap load assessment by aggregating features from sparse, handheld views of the pile. 
On 600 out-of-fold piles, frozen DINOv2 tokens aggregated uniformly across views reach 13.4\%
total variation (TV) distance against operator estimates.
Most remaining error lies in allocation among four thickness-defined heavy melting scrap (HMS) classes, 
yet image-dependent allocation reduces within-HMS TV from 25.1\% under a training-set split prior to 11.2\%.
Due to the lack of public datasets and benchmarks for predicting composition, 
external evaluation relies on the public DOES dataset intended for classification. 
A classifier using the same frozen features reaches 93.97\% accuracy, exceeding the published results we compare against. The composition model also aligns with related DOES categories without adaptation.
\end{abstract}

%% file: sec/v3_1_intro.tex
\section{Introduction}
\label{sec:introduction}

Increasing the recycling of steel scrap is important for reducing environmental impact and primary resource demand \cite{iea2020}. 
Electric-furnace steelmaking has increased its share of global crude-steel production from 25.1\% in 2015 to 30.3\% in 2025 \cite{worldsteel2016,worldsteel2026}. 
Scrap composition affects material valuation, downstream process planning, and expected yield, 
while scrap quality also influences the energy and environmental impact of Electric Arc Furnace (EAF) operation \cite{haupt2017,matino2017}. 
Despite its importance, composition assessment in industrial practice still commonly relies on trained human operators.

Computer vision has been explored for scrap classification, detection, and segmentation \cite{dossantos2025,xu2023,daigo2023,xiao2024}, 
with the broader goal of assisting or automating scrap assessment. Many of these approaches rely on object- or pixel-level annotations 
that are not produced during routine grading. 
Public datasets are also limited, and to the best of our knowledge, there is no public benchmark for mass-composition estimation.

\begin{figure*}[t]
\centering
\includegraphics[width=\textwidth]{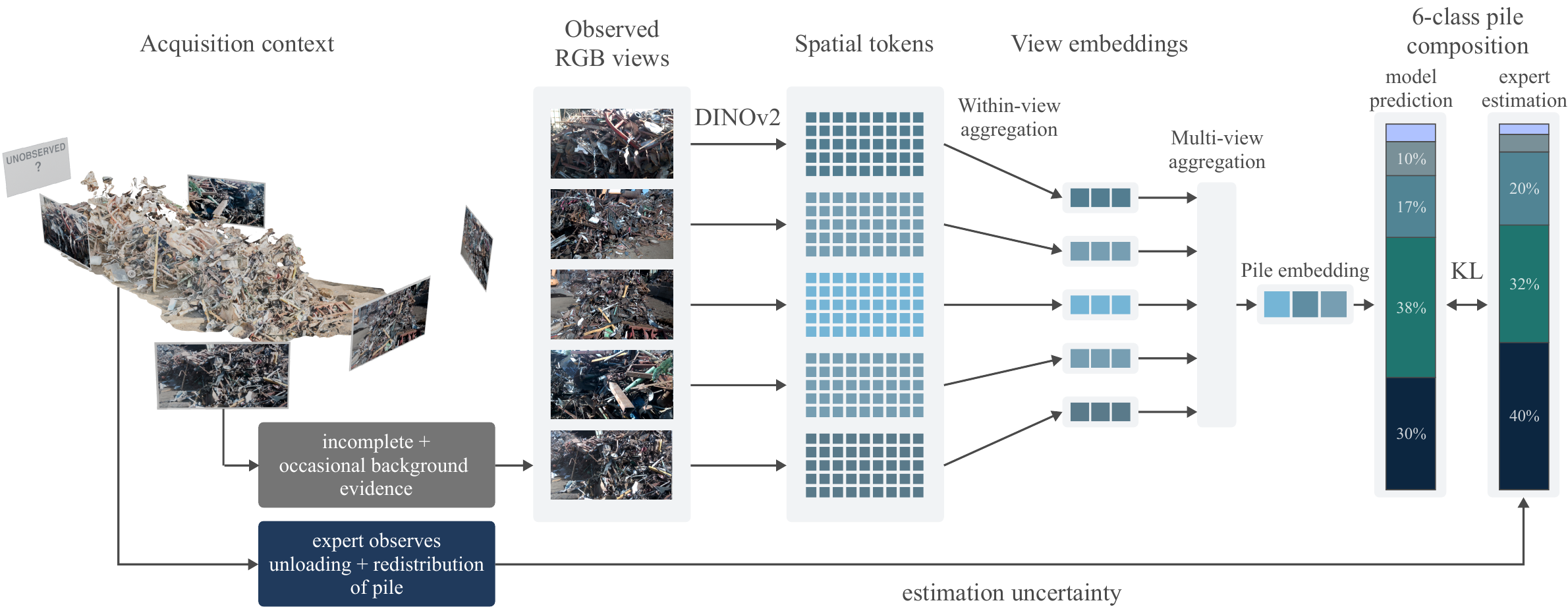}
\caption{Predicting a six-class operator estimate from a set of RGB
views. The operator also observes unloading and handling of the pile.}
\label{fig:task-overview}
\end{figure*}

LASER~\cite{lei2025laser} tackles composition estimation using human mass estimates,
rather than compositions obtained by physically sorting and weighing the load.
Related grading systems, such as Tu et al.~\cite{tu2022}, instead predict an
overall grade from segmented scrap proportions and counts.

Research on human perception provides another perspective on this setting. Visual material cues affect expectations 
about density and weight \cite{buckingham2009material}, while estimates of unfamiliar objects are influenced by learned 
relationships between size, density, and mass \cite{peters2015smaller}. More broadly, work on material perception and 
intuitive physics suggests that people combine visual observations with prior knowledge about materials and 
physical behavior \cite{xiao2025material,battaglia2013simulation}. 
This motivates asking how much of an operator estimate can be recovered from partial visual evidence alone.

We therefore study a simpler acquisition setting, which we call SteelHuman, based on a variable number of sparse RGB views captured with a handheld camera. 
Our main questions are how much of the operator's composition estimate can be recovered from sparse views, how the amount of visual evidence affects prediction 
and where the remaining errors are concentrated. 
We use frozen DINOv2 ViT-L/14 features with mean aggregation \cite{oquab2024,dosovitskiy2021}
(Fig.~\ref{fig:task-overview}) and compare learned aggregation, frozen backbones,
and inexpensive scores for identifying unreliable predictions.

Because the composition dataset is proprietary, we also use DOES to test transfer
to related public categories and supervised classification with the same frozen features.

Our contributions can be summarized as: (1) a low-burden formulation that predicts operator composition estimates from sparse photographs retained during routine grading, improving over non-visual priors; (2) an analysis of where the error lies, how it changes with the number of views, and how well inexpensive uncertainty scores flag unreliable piles, including a view-weighting oracle checked against a structured null; and (3) external checks on the public DOES dataset: transfer to related categories without adaptation, and a supervised classifier above the published results we compare against.

%% file: sec/v3_2_related_work.tex
\section{Related Work}
\label{sec:related-work}

\subsection{Vision-based scrap assessment}
Scrap inspection has progressed from hybrid sensing and handcrafted image
features \cite{kutila2005,wieczorek2008} to learned classification, detection,
and segmentation models \cite{dossantos2025,xu2023,xu2024rating,daigo2023,xiao2024}.
Xu et al.~\cite{xu2025mspfnet} address segmentation and classification
in scrap scenes.
DOES provides public European scrap-category images for classification \cite{schaefer2023does}, and its extension 
CLRiuS~\cite{schaefer2024clrius} studies self-supervised scrap representations on the same dataset.
Additional work on DOES evaluates efficient deployment on Jetson hardware by 
utilizing knowledge distillation \cite{esfandiari2026jetson}. 
These models predict categories or regions, which cannot be directly interpreted as mass compositions.

Tu et al.~\cite{tu2022} combine foreground extraction and scrap segmentation
in trailer images from static overhead cameras. Their Bayes-based classifier
uses the resulting proportions and counts to predict an overall grade, rather
than a mass-composition vector. LASER~\cite{lei2025laser} directly addresses
composition estimation and processes
truck-unloading video using key-frame selection, scene-change detection,
instance segmentation, and mask-aware recognition, then combines local
estimates using volume- and density-related information.
Storonkin et al.~\cite{storonkin2026} use unloading imagery
and MIL to predict inspector estimates of non-metallic contamination at the railcar level. 
All three use unloading imagery and controlled acquisition at designated sites, while Tu et al. and LASER~\cite{tu2022,lei2025laser} also rely on multi-stage preprocessing.
Our task predicts a six-class composition using a simpler
acquisition protocol: sparse handheld photographs from accessible viewpoints,
without requiring complete coverage of the pile.

Other studies introduce physical structure or richer supervision.
Díaz-Romero et al.~\cite{diazromero2022} estimate individual scrap-object mass
using depth-derived geometric features. Ji et al. and Yin et al.~\cite{ji2025,yin2026} both use segmentation to predict weight. 
Quan et al.~\cite{quan2026} use prepared
steel/copper mixtures with measured weights and segmentation annotations to
classify copper weight-percentage ranges. Gursch et al.~\cite{gursch2026}
investigate hyperspectral scrap characterization for composition optimization.
These settings illustrate the distinction between controlled environment measurements and operational
grading estimates.

\subsection{Predicting latent quantities and set supervision}
Visual mass inference is underconstrained because appearance does not
uniquely specify volume or density. Image2mass~\cite{standley2017image2mass}
studies appearance and object dimensions for inferring mass. Visual estimation of
food quantity and nutritional content offers another example of inferring latent
properties from images \cite{meyers2015,thames2021}. These tasks also require inferring quantities that are only partly visible.

Our variable-size, unordered views require a permutation-invariant set model \cite{zaheer2017}, 
with Multiple Instance Learning (MIL) providing a
natural learned-weight alternative \cite{ilse2018}. 
Although the labels give mass proportion, the supervision is not
ordinary Learning from Label Proportions (LLP) \cite{quadrianto2009,dulacarnold2019}:
views are repeated observations of one pile, rather than disjoint material
instances whose class counts define the target. MIL is therefore an aggregation framework here, while the target is a pile-level mass estimate rather than an instance-count proportion.
Multi-view shape recognition likewise combines observations across sparse viewpoints
\cite{su2015mvcnn,wei2021canonical}. 

%% file: sec/v3_3_operational_task.tex
\section{Task Formulation and Dataset}
\label{sec:operational-task}

\subsection{Operational process}
A truck unloads its contents to form a pile while an operator observes unloading and subsequent
handling. During this, the operator takes photographs with a handheld tablet from accessible viewpoints, ultimately recording
one composition estimate at one percentage point resolution, after the pile has been handled
(Fig.~\ref{fig:task-overview}). Each pile is graded by a single operator. 
Images can range from close-ups to overviews and
additional views may be retained when handling exposes informative material.

For pile $i$, let $X_i^\star\in\Delta^5$ denote its unobserved physical mass
composition. The available input and supervision are
\begin{equation}
E_i=\{I_{iv}\}_{v=1}^{V_i},\quad
\mathbf y_i=X_{\mathrm{op},i}\in\Delta^5,\quad
\hat{\mathbf y}_i=q_\theta(E_i).
\label{eq:target}
\end{equation}
The retained views and the recorded estimate are two imperfect observations of the same pile:
views can omit, occlude or repeat material, while the operator sees more than the photographs and brings personal judgment to the estimate. 
View count is not controlled during acquisition.

\subsection{Dataset}
Our dataset was collected at one plant and contains grades from eight operators.
We study the six plant categories that can be found in heterogeneous piles: Stamped steel, Cast iron, and heavy melting scrap (HMS) in four thickness grades (Table~\ref{tab:dataset}). Stamped steel and Cast iron are defined by product and material, whereas HMS is defined by thickness, although operators may also weigh thickness when separating thick Stamped steel from $>6$ mm HMS.
The dataset contains 750 piles: 223 with homogeneous targets and 527 with heterogeneous targets, with a median of 5 views per pile (range 1–19) and a total of 4,187 views.
Homogeneous targets contain one class while heterogeneous targets contain two or more.
Table~\ref{tab:dataset} gives the class distribution. 
The three thinner HMS classes occur only in heterogeneous piles, while Stamped steel, $>6$ mm HMS, and Cast iron also have homogeneous samples. 
The dataset is proprietary and will therefore not be released.

\subsection{Development and split stratification}
Five fixed, pile-disjoint folds contain 150 piles each. Stratification across folds uses the dominant target class 
and active-class count. Folds are not held out by date or operator.
$F_0$ is the development fold:
candidate configurations are trained on the other 600 piles and evaluated on
$F_0$. Primary results pool held-out predictions from $F_1$--$F_4$ after fixing
the training configuration. Each held-out-fold model is trained on the other four folds. This is a non-nested evaluation because $F_1$--$F_4$ also contribute
training data to development-fold comparisons.

\begin{table}[t]
\centering
\caption{Dataset class distribution. Active counts include heterogeneous piles.
Mean shares include zeros. Active: piles with a nonzero target share.}
\label{tab:dataset}
\small
\setlength{\tabcolsep}{3pt}
\input{tables/refreshed/class_rows}
\begin{tabular*}{\columnwidth}{@{\extracolsep{\fill}}lrrr@{}}
\toprule
Class & Active & Mean share (\%) & Homogeneous \\
\midrule
\RefreshedClassRows
\bottomrule
\end{tabular*}
\end{table}

%% file: tables/refreshed/class_rows.tex
\newcommand{\RefreshedClassRows}{%
Stamped & 99 & 13.0 & 96 \\
HMS $>6$ mm & 596 & 49.6 & 118 \\
HMS 4--6 mm & 506 & 15.1 & 0 \\
HMS 2--4 mm & 465 & 16.6 & 0 \\
HMS 0--2 mm & 251 & 3.3 & 0 \\
Cast iron & 365 & 2.4 & 9 \\
}

%% file: sec/v3_4_model.tex
\section{Model}
\label{sec:model}

\subsection{Mean pooling}
Our main model uses frozen DINOv2 ViT-L/14 features, extracted with automatic mixed precision.
The $3840\times2160$ or $1920\times1080$ images are rotated by their EXIF orientation, converted to RGB, and
resized directly to $896\times504$ with bicubic interpolation,
without cropping or padding. They are normalized using the DINOv2/ImageNet
channel statistics. This gives a $64\times36$ token grid, which $4\times4$ average pooling reduces to 144 spatial tokens.
Each token is then projected through a shared linear $1024\to256$
layer, LayerNorm, GELU, and dropout $p=0.1$. We average projected tokens
within each view and then average view embeddings. With equal token counts per
view, this is equivalent to averaging all tokens across views.
Finally, the output is a linear six-class head with a
softmax activation. The no-projector ablation instead feeds
mean-pooled frozen features directly to a six-class linear head.

\subsection{Learned aggregation}
The gated attention-based multiple-instance learning (ABMIL) \cite{ilse2018}
averages the projected spatial tokens within a view and learns
normalized weights over view embeddings using an attention dimension of 64 and no
attention dropout. Our Self-attention \cite{vaswani2017} variant is inspired by VGGT's alternating global/frame attention
\cite{wang2025vggt}. It applies one self-attention layer across all tokens
and one independently within each view, with four heads, feed-forward width
256, and dropout 0.1. No spatial or view positional embeddings are used.
The resulting tokens are mean pooled before the same composition head.

\subsection{Training}
Training uses Kullback--Leibler (KL) divergence $D_{\mathrm{KL}}(\mathbf y\|\hat{\mathbf y})$,
AdamW~\cite{loshchilov2019adamw} with weight decay 0.05, and a batch size of two piles for 40 epochs. Learning rates warm up
over 10\% of optimization steps and decay by a cosine schedule to 1\% of the
peak. The peaks are $8\times10^{-4}$ for Mean pooling and Gated ABMIL and
$5\times10^{-5}$ for Self-attention. Training uses full precision, with gradient norms clipped to 1.0.
All composition evaluations use the final epoch checkpoint, without early stopping.
The main comparison uses initialization seeds 42, 156, and 2026.

Each training epoch samples eight views per pile uniformly without replacement,
or uses all views when fewer than eight are available. Mean pooling and Gated ABMIL use all
144 spatial tokens per sampled view, and Self-attention uniformly samples 30 unique tokens.
Each aggregator uses its best tested $F_0$ configuration, so the comparison
uses separately selected hyperparameters. The ordinal-loss control is described in Supplementary Sec.~A.

%% file: sec/v3_5_evaluation.tex
\section{Evaluation}
\label{sec:evaluation}

\subsection{Metrics}
Our primary metric is total variation distance,
\begin{equation}
\TV(\mathbf y,\hat{\mathbf y})=\frac12\sum_{c=1}^{6}|y_c-\hat y_c|.
\label{eq:tv}
\end{equation}
TV represents the minimum share of the composition vector that must be reassigned to match the operator estimate.
A non-visual baseline fits one composition per training split by minimizing mean TV.
For each initialization seed, we pool
the held-out $F_1$--$F_4$ predictions and compute each summary over piles.
Unless stated otherwise, evaluation uses all retained views and spatial tokens
with dropout disabled. Supplementary Table~S1 lists the evaluation cohorts.

To compare with LASER \cite{lei2025laser}, let
$q_{ic}=y_{ic}/\sum_{d=1}^{4}y_{id}$ and
$\hat q_{ic}=\hat y_{ic}/\sum_{d=1}^{4}\hat y_{id}$ for the four HMS classes,
indexed here by $c=1,\ldots,4$. The pooled RMSE is
\begin{equation}
\mathrm{RMSE}_{\mathrm{HMS}}
=\sqrt{\frac{1}{4N}\sum_{i=1}^{N}\sum_{c=1}^{4}
(\hat q_{ic}-q_{ic})^2}.
\label{eq:rmse-hms}
\end{equation}
This comparison uses the 405 heterogeneous piles with at least 95\% target HMS.
The threshold retains mostly HMS piles with small Cast iron shares (1--5\%)
that an HMS-only restriction would exclude.

\subsection{Error structure and additional diagnostics}
\begin{figure*}[t]
\centering
\includegraphics[width=\textwidth]{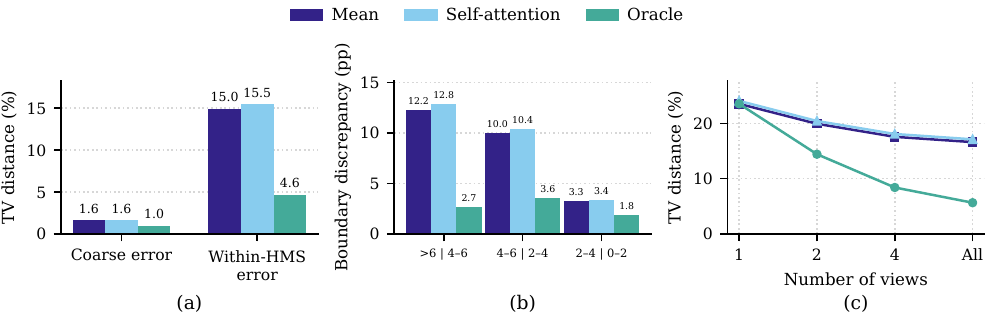}
\caption{Seed-42 models on 356 heterogeneous $F_1$--$F_4$ piles with
at least four views. (a) All-view coarse and within-HMS TV.
(b) Mean absolute differences in cumulative HMS-normalized shares at the 6, 4, and
2\,mm boundaries, in percentage points.
(c) TV with controlled view counts, averaging 30 nested subsets per pile
for each fixed count. Oracle weights use
the recorded target and are optimized separately for each subset.}
\label{fig:error-view-diagnostics}
\end{figure*}

\paragraph{Coarse and ordinal error.}
Collapsing the four HMS classes gives Stamped steel/HMS/Cast iron compositions.
We report coarse TV and the nonnegative residual
$\TV-\TV_{\mathrm{coarse}}$, termed within-HMS redistribution TV.
For target HMS mass $m_i\geq0.10$, we independently normalize HMS targets and
predictions and compute the 1-Wasserstein distance over ordered thickness classes
\begin{equation}
W_{1,i}=\sum_{j=1}^{3}
\left|\sum_{c=1}^{j}(q_{ic}-\hat q_{ic})\right|.
\label{eq:w1}
\end{equation}
Adjacent thickness classes have unit distance: $W_1$ is in class steps, not
millimetres. Its three parts are cumulative discrepancies across the
6, 4, and 2\,mm boundaries. Table~\ref{tab:hms-prior} reports unweighted means of conditional TV and $W_1$ over eligible piles. 
The aggregate displacement per unit of redistribution is
\begin{equation}
\bar d_{\mathrm{agg}}=\frac{\sum_i W_{1,i}}{\sum_i\TV(q_i,\hat q_i)}.
\label{eq:aggregate-distance}
\end{equation}
Figure~\ref{fig:error-view-diagnostics}b shows these cumulative boundary
discrepancies on the view-count cohort.

\paragraph{Within-HMS prior control.}
For each held-out fold, we average the normalized four-class HMS targets
of training piles with positive HMS mass. We keep each model prediction's Stamped steel
share, Cast iron share, and total HMS share, but divide its HMS share using
this fixed training prior. This tests whether image-dependent HMS allocations
improve on a constant split while keeping the coarse prediction unchanged.

\paragraph{View subset evaluation.}
For the seed-42 Mean pooling and Self-attention models, we evaluate the same nested subsets with
$k\in\{1,2,4,\mathrm{all}\}$ over 30 fixed repetitions on the same 356
heterogeneous $F_1$--$F_4$ piles having at least four views. We first average
subset errors across repetitions within each pile.

\begin{table*}[t]
\centering
\caption{Primary out-of-fold TV (\%). For trained models, each metric is
computed across piles within each seed and reported as mean $\pm$ SD across
three seeds. The prior has no training-seed variation. Bold marks the lowest
observed mean.}
\label{tab:main-results}
\small
\setlength{\tabcolsep}{3pt}
\input{tables/advisor/aggregation_rows}
\begin{tabular*}{\textwidth}{@{\extracolsep{\fill}}lcccccc@{}}
\toprule
& \multicolumn{2}{c}{All piles ($N=600$)}
& \multicolumn{2}{c}{Homogeneous ($N=180$)}
& \multicolumn{2}{c}{Heterogeneous ($N=420$)} \\
\cmidrule(lr){2-3}\cmidrule(lr){4-5}\cmidrule(l){6-7}
Method & Mean $\downarrow$ & Median $\downarrow$
& Mean $\downarrow$ & Median $\downarrow$ & Mean $\downarrow$ & Median $\downarrow$ \\
\midrule
\AdvisorAggregationRows
\bottomrule
\end{tabular*}
\end{table*}

\paragraph{Target-conditioned view oracle.}
On these same $F_1$--$F_4$ subsets, the target-conditioned oracle optimizes
view weights separately for each subset while keeping the Mean pooling model frozen.
Softmax weights combine projected
view embeddings, $h=\sum_v a_vh_v$, before the composition head.
A separate $F_0$ diagnostic uses the 102 of 107 heterogeneous piles for which other targets satisfying the matching constraints exist, 
the frozen seed-42 Mean pooling model, and 30 matched
nested subsets. If fewer than $k$ views exist in this diagnostic, all available
views are used. Optimization details are in Supplementary Sec.~B.

Five target shuffles among eligible heterogeneous $F_0$ piles preserve dominant class and active-class
count, constrain dominant-share differences to at most 10 percentage
points, and exclude the original target sample. Exact active classes
are not always preserved. Original and shuffled targets share the same pile
and candidate subsets. 

\subsection{Selective prediction and external evaluation}
\paragraph{Risk--coverage.}
On the 403 heterogeneous $F_1$--$F_4$ piles with at least two views, we compare
mean leave-one-view-out (LOVO) shift
$S=V^{-1}\sum_v\TV(\hat{\mathbf y}(E),\hat{\mathbf y}(E\setminus\{I_v\}))$,
normalized shift $(V-1)S$, full-view prediction entropy
$-\sum_c\hat y_c\log\hat y_c$, and view count.
LOVO measures how much the prediction changes when each view is removed.
Lower scores are retained first, except that larger view counts are retained
first. Risk is the mean TV of the retained piles (Fig.~\ref{fig:risk-coverage}).

\paragraph{DOES protocols.}
We distinguish two tasks. Semantic transfer evaluates our pile-trained models
without adaptation on four related DOES categories~\cite{schaefer2023does}: E1 maps to 0--6\,mm
HMS, E3 and EHRB (rebar) to $>6$\,mm HMS and E8 to thin
new-production scrap matching our Stamped steel.
The composition models are trained only
on our proprietary pile data. Their outputs on the full available DOES
collection for these categories are averaged across $F_0$--$F_4$ models. 
This experiment uses the ``raw'' set of two overlapping large crops. 
We upsample each crop to $504\times504$ and extract DINOv2 ViT-L tokens.
Local $4\times4$ averaging gives 81 tokens per crop, which are projected
before averaging within and across crops. We average model outputs after softmax.

Supervised DOES classification uses the official 105,523 training and 8,131
temporally separate test tiles. Test targets cover eight scrap classes and the
ninth output, background, has no test examples but remains eligible during prediction.
We use all 256 spatial DINOv2-L tokens from $224\times224$ tiles, without local token pooling.
Tokens are projected when a projector is used, mean pooled, and passed to a linear classification head.
We compare a new linear head directly on the frozen features, a head on the
frozen pile-trained projector, and joint head/projector training from either
pile-trained or random projector initialization. The pretrained projector is taken from
the Mean pooling model trained with $F_0$ held out. All four use initialization seed 2026,
20 fixed epochs, cross-entropy loss, AdamW, learning rate $3\times10^{-4}$, weight decay 0.01,
batch size 128, two warmup epochs and cosine decay, without augmentation.

\paragraph{Centered token contributions.}
For projected token $\mathbf z_t$ and class-head weights $\mathbf w_c$, we
display $(\mathbf w_c-\bar{\mathbf w})^\top\mathbf z_t$, with
$\bar{\mathbf w}=\frac16\sum_c\mathbf w_c$.
Averaging these contributions and adding the centered head bias recovers
the centered pile logits, with unchanged softmax probabilities.
Supplementary Fig.~S3 shows example maps.

%% file: tables/advisor/aggregation_rows.tex
\newcommand{\AdvisorAggregationRows}{%
TV-optimal prior & 40.44 & 36.00 & 67.76 & 39.00 & 28.73 & 26.00 \\
\midrule
Mean pooling & $\mathbf{13.44\pm0.03}$ & $\mathbf{10.20\pm0.14}$ & $5.05\pm0.04$ & $0.20\pm0.03$ & $\mathbf{17.03\pm0.03}$ & $\mathbf{15.04\pm0.22}$ \\
Self-attention & $13.76\pm0.35$ & $10.40\pm0.24$ & $\mathbf{4.60\pm0.24}$ & $\mathbf{0.03\pm0.00}$ & $17.69\pm0.43$ & $15.15\pm0.10$ \\
Gated ABMIL & $15.43\pm0.11$ & $11.84\pm0.28$ & $6.07\pm0.10$ & $0.12\pm0.01$ & $19.45\pm0.11$ & $16.30\pm0.18$ \\
Mean pooling, no projector & $14.03\pm0.02$ & $11.16\pm0.21$ & $6.87\pm0.10$ & $0.53\pm0.02$ & $17.10\pm0.06$ & $15.05\pm0.11$ \\
}

%% file: sec/v3_6_results.tex
\section{Results}
\label{sec:results}

\subsection{Aggregation and backbones}
Table~\ref{tab:main-results} reports the three-seed comparison.
Mean pooling has the lowest overall and heterogeneous TV. Self-attention is better on homogeneous piles.
Paired comparisons are given in Supplementary Table~S2. 
The no-projector ablation increases overall TV by 0.59 percentage points.
The small heterogeneous-pile change suggests that frozen features already provide much
of the information used by the trained projector.
Median TV shows that predictions for most homogeneous targets are nearly exact,
with a few severe failures increasing the mean TV (Supplementary Fig.~S2).

To check the Self-attention train--test token-count difference, we also
evaluate 30 randomly sampled tokens per view, keeping all views and averaging
predictions over 20 draws: seed-42 TV is 13.80\%, versus 13.69\% with all
144 tokens. Matching the training token count therefore does not improve this comparison
(see Supplementary Sec.~A for individual-draw results).

\begin{table}[t]
\centering
\caption{Backbone comparison on the development fold ($N=150$).
TV (\%) from one run per backbone using Mean pooling.}
\label{tab:backbone-results}
\small
\setlength{\tabcolsep}{3pt}
\begin{tabular*}{\columnwidth}{@{\extracolsep{\fill}}lrrr@{}}
\toprule
Representation & All & Hom. & Hetero. \\
\midrule
DINOv2 ViT-S \cite{oquab2024} & 14.72 & 10.56 & 16.39 \\
DINOv2 ViT-L \cite{oquab2024} & 12.83 & 5.15 & 15.92 \\
DINOv2 ViT-g \cite{oquab2024} & 12.84 & 5.67 & 15.72 \\
DINOv3 ViT-L \cite{simeoni2025dinov3} & 13.30 & 6.04 & 16.22 \\
ConvNeXt-V2-L \cite{woo2023convnextv2} & 15.61 & 9.31 & 18.14 \\
CLIP ViT-L \cite{clip} & 15.87 & 11.62 & 17.58 \\
\bottomrule
\end{tabular*}
\end{table}

Table~\ref{tab:backbone-results} reports the $F_0$ backbone comparison.
Final-layer DINOv2 ViT-L and ViT-g have nearly identical TV and the lowest overall error in this comparison.
Backbone extraction details are in Supplementary Sec.~E.
Intermediate layers and learned layer fusion do not improve overall TV over
the final layer (Supplementary Table~S3).

\subsection{HMS error}
On all 600 primary piles, seed-42 coarse TV is 2.24\%, and 83.3\% of
total TV remains within HMS. Figure~\ref{fig:error-view-diagnostics} shows
the same decomposition and cumulative boundary discrepancies on the separate
356-pile view-count cohort. There, the largest discrepancy is at 6\,mm for
Mean pooling and Self-attention, and at 4\,mm for the oracle.

\paragraph{Visual HMS allocation.}
Replacing the predicted HMS split with the training prior substantially
increases within-HMS TV from 11.19\% to 25.12\% on the 600 primary piles
(Table~\ref{tab:hms-prior}). The model therefore predicts
pile-specific HMS allocations beyond the average training composition.
Among the 515 piles with at least 10\% target HMS, aggregate displacement
is 1.61 class steps for the model and 1.74 for the prior,
so errors extend beyond adjacent classes even though the model substantially
improves allocation. These comparisons concern operator estimates, not measured thickness.
Supplementary Sec.~A checks alternative priors and piles with at least two active HMS classes.
Adding an ordinal $W_1$ penalty with a weight of 0.1 left performance essentially
unchanged in the matched seed-42 experiment.

On the restricted 405-pile cohort, our Mean pooling model reaches an RMSE of 11.8\%, 
close to the 11.9\% LASER reports on its own data. Because of the different dataset and class taxonomy,
this can only be considered a scale reference rather than a direct comparison.

\subsection{View count and target-conditioned gains}
Additional views reduce Mean pooling TV from 23.5\% with one view to 16.6\% with all views (Fig.~\ref{fig:error-view-diagnostics}c). 
The all-view endpoint uses
a variable number of views per pile, so its gain is not a fixed per-view effect.
On this cohort, target-conditioned oracle weighting reduces all-view TV
from 16.6\% to 5.7\% (Fig.~\ref{fig:error-view-diagnostics}c).
The separate $F_0$ control tests whether such gains are specific to the recorded target.

On the 102-pile $F_0$ control, recorded and shuffled targets yield similar
all-view gains (9.3 and 9.7 percentage points). Recorded targets have lower
TV both before reweighting (14.7\% versus 17.1\%) and after it (5.3\% versus
7.4\%). Thus several structurally similar targets can benefit from reweighting
the same views (Supplementary Fig.~S1).

\begin{table}[t]
\centering
\caption{Within-HMS prior control, seed 42, $F_1$--$F_4$.
The prior retains the model's coarse prediction. Redistribution uses all
600 piles. Conditional metrics use the 515 with at least 10\% target HMS.}
\label{tab:hms-prior}
\small
\setlength{\tabcolsep}{3pt}
\begin{tabular*}{\columnwidth}{@{\extracolsep{\fill}}lrr@{}}
\toprule
Metric $\downarrow$ & Model & HMS prior \\
\midrule
Within-HMS TV (\%) & 11.19 & 25.12 \\
Conditional HMS TV (\%) & 14.13 & 30.74 \\
Conditional $W_1$ (class steps) & 0.227 & 0.534 \\
\bottomrule
\end{tabular*}
\end{table}

\subsection{External evaluation on DOES}
\begin{figure}[t]
\centering
\includegraphics[width=\columnwidth]{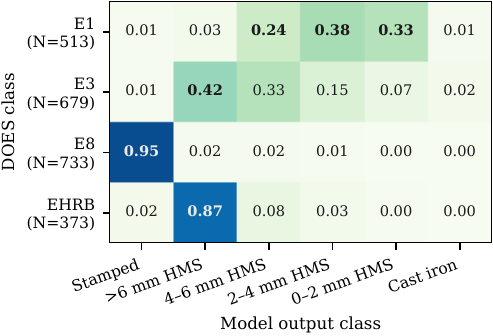}
\caption{External transfer without DOES training. Cells show mean composition
outputs by DOES category across the $F_0$--$F_4$ models. $N$ is the number
of source-image crop pairs. Bold entries mark predefined category correspondences.}
\label{fig:does}
\end{figure}

Without DOES training (Fig.~\ref{fig:does}), E1 emphasizes thinner HMS,
EHRB favors $>6$\,mm HMS, and E8 favors Stamped steel. E3 is less clearly
separated: 42\% goes to $>6$\,mm HMS and 33\% to 4--6\,mm HMS,
with 55\% across the three thinner HMS classes including 4--6\,mm. 

Training the pile-initialized projector and a new head reaches 93.97\%,
1.17 percentage points above the highest published accuracy listed in
Table~\ref{tab:does-classification}. Random projector initialization gives
93.91\%, so pile initialization adds little. Frozen DINO features also perform better than the pile-pretrained frozen projector.

\subsection{Selective prediction and deployment}

Prediction entropy performs similarly to LOVO without repeated prediction:
at 50\% coverage, retained TV is about 15.0\% for both, versus 16.7\%
under random retention (Fig.~\ref{fig:risk-coverage}). This offers a single forward pass for triage, 
although even the retained half has substantial error. Supplementary Fig.~S5
shows the pooled analysis including homogeneous piles. 
Ranking by view count is worse than random, which could point to operators keeping more views for harder piles.

\begin{table}[t]
\centering
\caption{Classification on the official DOES test tiles ($N=8{,}131$).}
\label{tab:does-classification}
\small
\setlength{\tabcolsep}{2pt}
\input{tables/refreshed/does_ours_rows}
\begin{tabular*}{\columnwidth}{@{\extracolsep{\fill}}lr@{}}
\toprule
DOES-trained method & Accuracy (\%) \\
\midrule
Original DOES, PreActResNet18 \cite{schaefer2023does} & 68.09 \\
CLRiuS, 10 labeled tiles/class \cite{schaefer2024clrius} & 75.89 \\
LASER, pretrained CLIP \cite{lei2025laser} & 92.80 \\
Esfandiari et al.~\cite{esfandiari2026jetson}, ViT-L teacher & 92.37 \\
Esfandiari et al.~\cite{esfandiari2026jetson}, distilled DeiT & 88.81 \\
\midrule
\RefreshedDoesRows
\bottomrule
\end{tabular*}
\end{table}

We profile the Mean pooling $F_0$ model on a GTX 1080 Ti using FP16 feature
extraction and FP32 aggregation. The benchmark uses 50 piles covering the whole view-count range and a batch size of one.
Mean latency is 2.06\,s per pile, or 382\,ms per view, with DINO feature extraction
taking 99.7\% of the time.
These timings exclude loading, warmup, disk preload,
networking, queues, and serialization, so tablet-to-server overhead is not included.

%% file: tables/refreshed/does_ours_rows.tex
\newcommand{\RefreshedDoesRows}{%
Ours: frozen DINOv2-L, linear head & 91.53 \\
Ours: frozen pile projector & 85.71 \\
Ours: train random projector & 93.91 \\
Ours: train pile-initialized projector & 93.97 \\
}

%% file: sec/v3_7_discussion.tex
\section{Discussion}
\label{sec:discussion}

Simple acquisition through sparse handheld views shows promising performance when trained on human composition estimates.
The main difficulty is allocating mass among HMS thickness classes.
The prior controls show that the model learns pile-specific allocations even for the harder HMS cases,
but do not establish which cues it uses: appearance properties such as thickness, texture, and shape could all correlate with operator estimates. 
The lack of homogeneous targets for the three
thinner HMS classes also confounds the comparison. 
A possible explanation for the lack of homogeneous thinner samples is that
Cast iron, Stamped steel and thick metal pieces such as beams often come from one place and are easier to sort due to their material definition, while sorting thinner steel is laborious, which is one of the reasons why this process still relies on human estimates. 
This could contribute to the worse performance on thinner classes, whose boundaries are also harder to judge visually.

\begin{figure}[t]
\centering
\includegraphics[width=\columnwidth]{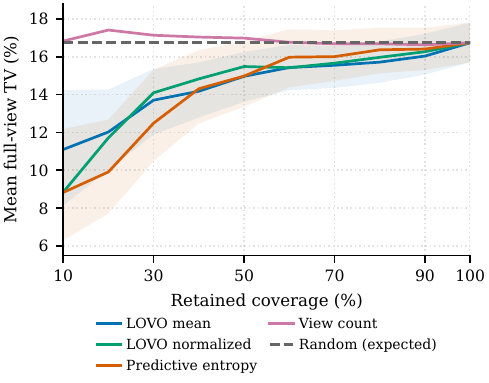}
\caption{Risk--coverage for seed-42 Mean pooling on 403 heterogeneous $F_1$--$F_4$
piles with at least two views. Risk is the retained mean full-view TV, and the dashed line shows expected risk under random retention. Shading denotes pointwise 95\% pile-bootstrap
intervals for LOVO mean and predictive entropy.}
\label{fig:risk-coverage}
\end{figure}

Low mean coarse error does not eliminate severe category mistakes.
Homogeneous targets account for about 48\% of the total coarse error in
the seed-42 evaluation of 600 primary piles. The five homogeneous piles with TV above
90\% account for about 29\%. Some piles graded as $>6$\,mm HMS visually resemble
Stamped steel and are predicted as such (Supplementary Fig.~S2). Thickness
and the operator's taxonomy may affect this distinction. Homogeneous HMS
targets can also incur within-HMS errors when predicted as thinner classes.

The retained views do not capture all the evidence used by the operator:
views repeat or omit material, and operator estimates may vary.
The structured null preserves dominant class and active-class count while keeping
dominant shares close, so similar reweighting gains for these nearby targets are plausible.
The oracle gap is a diagnostic of the representation, not an estimate of what a learned aggregator could recover.
Without independent grades or measured fractions, we cannot separate model error
from target ambiguity.

The no-projector ablation and DOES classification both point to a strong
contribution from frozen features. The tested learned aggregators do not
improve overall TV over mean pooling, despite modeling differences or interactions among views. 
Richer aggregation may help, with more training data, exhaustive evidence and measured targets making those relationships learnable.

One of the advantages of mean pooling regarding interpretability is the token contribution decomposition
(Supplementary Fig.~S3).
These maps show regional contributions to the model's logits and could help
operators inspect a prediction.
Prediction entropy can help rank estimates for review (Fig.~\ref{fig:risk-coverage}),
but may miss severe homogeneous errors when the model is confidently wrong.
A deployment advantage of mean pooling is caching the pooled embeddings and allowing the
predictions to update as photographs arrive. 
Practical usefulness still requires deploying the system and recording how these outputs affect grading.

Further progress may require better evidence as well as better models.
Physical inspection can provide thickness information beyond appearance
(Supplementary Fig.~S4). Scale references or depth could also help, but sparse,
insufficiently overlapping or changing pile views need not support reliable
geometric reconstruction. Even surface reconstruction leaves buried contents unseen. A synthetic-pile dataset built from scans of
individual scrap pieces could offer controlled compositions and viewpoints,
provided the pieces' masses are measured. 
Repeated independent grades or measured subsets would also help determine
which disagreements can reasonably be resolved from photographs.

%% file: sec/v3_8_conclusion.tex
\section{Conclusion}
\label{sec:conclusion}

We asked how much of an operator's estimate can be recovered from the handheld photographs retained during routine grading. 
Frozen DINOv2 features with mean pooling recover a substantial part of it, 
and none of the learned aggregators we tested improved its overall TV. 
Coarse composition error is low, with most remaining disagreement concerning the split
among HMS thickness classes, where the model clearly improves on a constant training-set split.
These results motivate evaluating sparse handheld views for operator assistance.
Progress towards more autonomous grading will require more data, better reference targets,
and evaluation under deployment conditions.

%% file: sec/v3_code_availability.tex
\paragraph{Code availability.}
Code for the DOES experiments will be released at
\url{https://github.com/jovankamchev/steelhuman}.
The industrial dataset is proprietary and cannot be released.

%% file: sec/v3_9_acknowledgments.tex
\section*{Acknowledgments}
We thank Makstil AD Skopje for scholarship funding and for providing the
industrial data used in this study.

Large language models were used for coding (models, result compilation) and language editing. The authors take
responsibility for the research, analyses, and manuscript.

%% file: supplementary_sections/revised_protocols.tex
\section{Evaluation Details and Additional Controls}
\label{sec:supp-protocols}

Table~\ref{tab:supp-cohorts} lists the subsets used in the main paper.
Homogeneous targets have one active class; heterogeneous targets have two or more.
Primary comparisons pool held-out predictions from $F_1$--$F_4$.
The development-fold oracle control uses a separate cohort.

\begin{table}[H]
\centering
\caption{Evaluation cohorts. Unless marked $F_0$, all use $F_1$--$F_4$.
Main-paper Table~2 uses three seeds; diagnostics use seed 42.}
\label{tab:supp-cohorts}
\small
\setlength{\tabcolsep}{3pt}
\begin{tabular*}{0.85\textwidth}{@{\extracolsep{\fill}}lr@{}}
\toprule
Analysis / inclusion rule & Piles \\
\midrule
Primary evaluation & 600 \\
\quad Homogeneous & 180 \\
\quad Heterogeneous & 420 \\
Conditional HMS: target HMS $\geq10\%$ & 515 \\
Four-HMS RMSE: heterogeneous, HMS $\geq95\%$ & 405 \\
View subsets: heterogeneous, $V\geq4$ & 356 \\
Main risk--coverage: heterogeneous, $V\geq2$ & 403 \\
Pooled risk--coverage: all, $V\geq2$ & 567 \\
Structured null: eligible heterogeneous $F_0$ piles & 102 \\
Latency: view-count-stratified $F_0$ sample & 50 \\
\bottomrule
\end{tabular*}
\end{table}

\paragraph{Paired aggregation comparisons.}
For each pile, we average the errors of its three initialization seeds,
then compute the difference between methods. We resample piles within
held-out folds 10,000 times, retaining the paired predictions
(Table~\ref{tab:supp-paired}). These intervals describe pile uncertainty
for the fitted models, rather than variation from retraining on new data.

\paragraph{Self-attention token-count control.}
We keep the seed-42 checkpoints and all available views fixed, disable
dropout, and sample 30 tokens per view for each of 20 draws. Predictions
are averaged after softmax. On the 420 heterogeneous piles, full-token TV is
17.68\%, compared with 17.74\% for the averaged prediction and 18.23\%
for the average individual-draw error. Thus the Monte Carlo average is
close to full-token evaluation, while a single smaller-token draw is less
accurate on average.

\paragraph{Within-HMS prior sensitivity.}
The main prior averages conditional HMS targets over training piles with
positive HMS mass. The 10\% threshold defines evaluation eligibility only.
An alternative prior normalizes the mean unnormalized HMS shares;
another minimizes training conditional TV over a constant probability
vector. Their seed-42 redistribution TV is 25.11\% and 24.67\%, respectively,
versus 25.12\% for the main prior and 11.19\% for the model.
Across the three seeds, model redistribution TV ranges from 11.19\% to
11.20\%, versus 25.10\% to 25.12\% for the main prior.

\pagebreak
The difference also persists among the 411 evaluation piles with at least
two active HMS classes and at least 10\% HMS: model/prior redistribution
TV is 15.62/26.77\%. On all 600 primary piles, the prior-minus-model
difference is 13.94 percentage points, with a paired within-fold
95\% pile-bootstrap interval of $[12.44,15.42]$ (10,000 replicates).
These comparisons keep the predicted coarse composition unchanged.

\begin{table}[H]
\centering
\caption{TV difference from Mean pooling, in percentage points, with paired
95\% bootstrap intervals. Positive values favor Mean pooling.}
\label{tab:supp-paired}
\small
\setlength{\tabcolsep}{2pt}
\begin{tabular*}{0.85\textwidth}{@{\extracolsep{\fill}}lcc@{}}
\toprule
Method & All piles & Heterogeneous \\
\midrule
Self-attention & $0.32\;[-0.08,0.72]$ & $0.66\;[0.19,1.12]$ \\
Gated ABMIL & $1.99\;[1.27,2.73]$ & $2.41\;[1.59,3.26]$ \\
No projector & $0.59\;[0.09,1.14]$ & $0.07\;[-0.41,0.53]$ \\
\bottomrule
\end{tabular*}
\end{table}

\paragraph{Layer depth and fusion.}
Table~\ref{tab:supp-layers} compares DINOv2 ViT-L intermediate features with
the final layer on $F_0$. Two-layer fusion uses one learned scalar:
$\alpha\mathbf x_{17}+(1-\alpha)\mathbf x_{\mathrm{final}}$, with
$\alpha=\operatorname{sigmoid}(a)$. Four-layer fusion uses learned softmax
weights over layers 4, 11, 17, and the final layer. These weights are shared
across tokens, views, and piles. Aligned features are combined before the
shared projector and mean pooling. Neither fusion improves overall TV over
the final-layer baseline.

\begin{table}[H]
\centering
\caption{DINOv2 ViT-L layer depth and fusion on $F_0$ ($N=150$).
TV (\%) from one run per configuration, using Mean pooling.}
\label{tab:supp-layers}
\small
\setlength{\tabcolsep}{2pt}
\begin{tabular*}{0.85\textwidth}{@{\extracolsep{\fill}}lrrr@{}}
\toprule
Features & All & Hom. & Hetero. \\
\midrule
Final layer & 12.83 & 5.15 & 15.92 \\
\midrule
\multicolumn{4}{@{}l}{\textit{DINOv2 ViT-L layers}} \\
Layer 4 & 23.47 & 30.17 & 20.78 \\
Layer 11 & 15.04 & 10.86 & 16.71 \\
Layer 17 & 14.23 & 10.07 & 15.90 \\
\midrule
\multicolumn{4}{@{}l}{\textit{DINOv2-L layer fusion}} \\
Scalar mixing: 17/final & 12.96 & 5.93 & 15.79 \\
Softmax mixing: 4/11/17/final & 13.30 & 7.18 & 15.76 \\
\bottomrule
\end{tabular*}
\end{table}

\paragraph{Ordinal-loss control.}
The extra training term is $\lambda m_i W_{1,i}/3$, where $m_i$ is target
HMS mass and $W_{1,i}$ compares the normalized HMS distributions.
It is zero for targets without HMS. With $\lambda=0.1$ and seed 42,
overall $F_1$--$F_4$ TV is 13.44\%, versus 13.43\% for KL alone.
This tests one weighting of the ordinal term.

%% file: supplementary_sections/revised_controls.tex
\clearpage
\section{Oracle Optimization and Structured Control}
\label{sec:supp-oracle-control}

\paragraph{Oracle optimization.}
The oracle keeps the trained Mean pooling model frozen and optimizes softmax
weights over view embeddings separately for each subset. Adam minimizes TV
to the supplied target for 300 steps at learning rate 0.05, starting from
one uniform and two random weight initializations. We also retain uniform
weighting, single-view solutions, and solutions from smaller nested subsets,
then report the lowest error among these candidates. This keeps any solution
available with fewer views, but does not guarantee a global optimum.
The weights form convex combinations of view embeddings; the final softmax
means that the output probabilities need not be convex combinations of
single-view predictions.

\begin{figure}[H]
\centering
\includegraphics[width=0.48\textwidth]{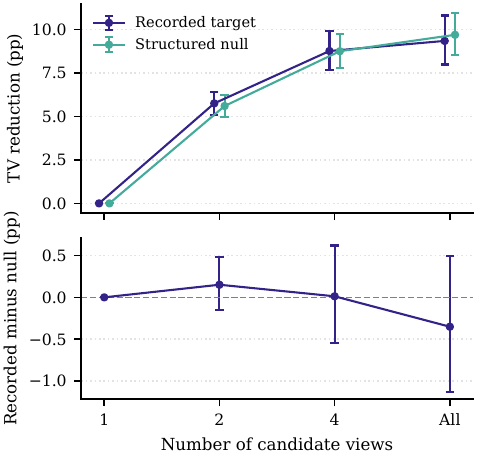}
\caption{Target-conditioned oracle on 102 heterogeneous $F_0$ piles.
Top: gain over uniform weighting. Bottom: paired recorded-minus-null gain.
Bars show 95\% pile-bootstrap intervals; candidate counts are capped at
the available views.}
\label{fig:supp-oracle-null}
\end{figure}

\paragraph{Structured-null oracle.}
Figure~\ref{fig:supp-oracle-null} compares target-conditioned gains on the
same 102 heterogeneous $F_0$ piles. Five shuffled targets per pile preserve the
dominant class and active-class count, with dominant shares within
10 percentage points; exact active-class identities may differ.
Recorded and shuffled targets use the same 30 nested view subsets.
Counts are capped at the available views: at $k=4$, 87 piles have four
candidates and 15 have fewer. With one view, the gain is zero by construction.

With all views, recorded/null oracle TV is 5.32/7.43\%, while the gains
over uniform weighting are 9.34/9.69 percentage points. Similar gains
therefore coexist with different residual errors. The result shows that
answer-informed reweighting can fit several structurally similar targets;
it does not rule out useful image information or learnable view weights.
Restricting the control to the 87 piles with at least four views gives recorded/null
gains of 10.07/10.50 percentage points and residual TV of 4.46/6.69\%.

%% file: supplementary_sections/homogeneous_failures.tex
\clearpage
\section{Large Disagreements on Homogeneous Piles}
\label{sec:supp-homogeneous-failures}

For a homogeneous target of class $c$, $\TV=1-\hat y_c$: a confident
prediction of another class can therefore produce nearly 100\% TV.
Among the 180 homogeneous primary piles, the seed-42 Mean model has
eight errors above 30\% TV, including five above 90\%.
The two largest errors are 99.99\% and 99.98\%. Both targets are entirely
$>6$\,mm HMS, whereas the model assigns 98.98\% and 99.97\%, respectively,
to Stamped steel.

\begin{figure}[H]
\centering
\includegraphics[width=0.66\textwidth]{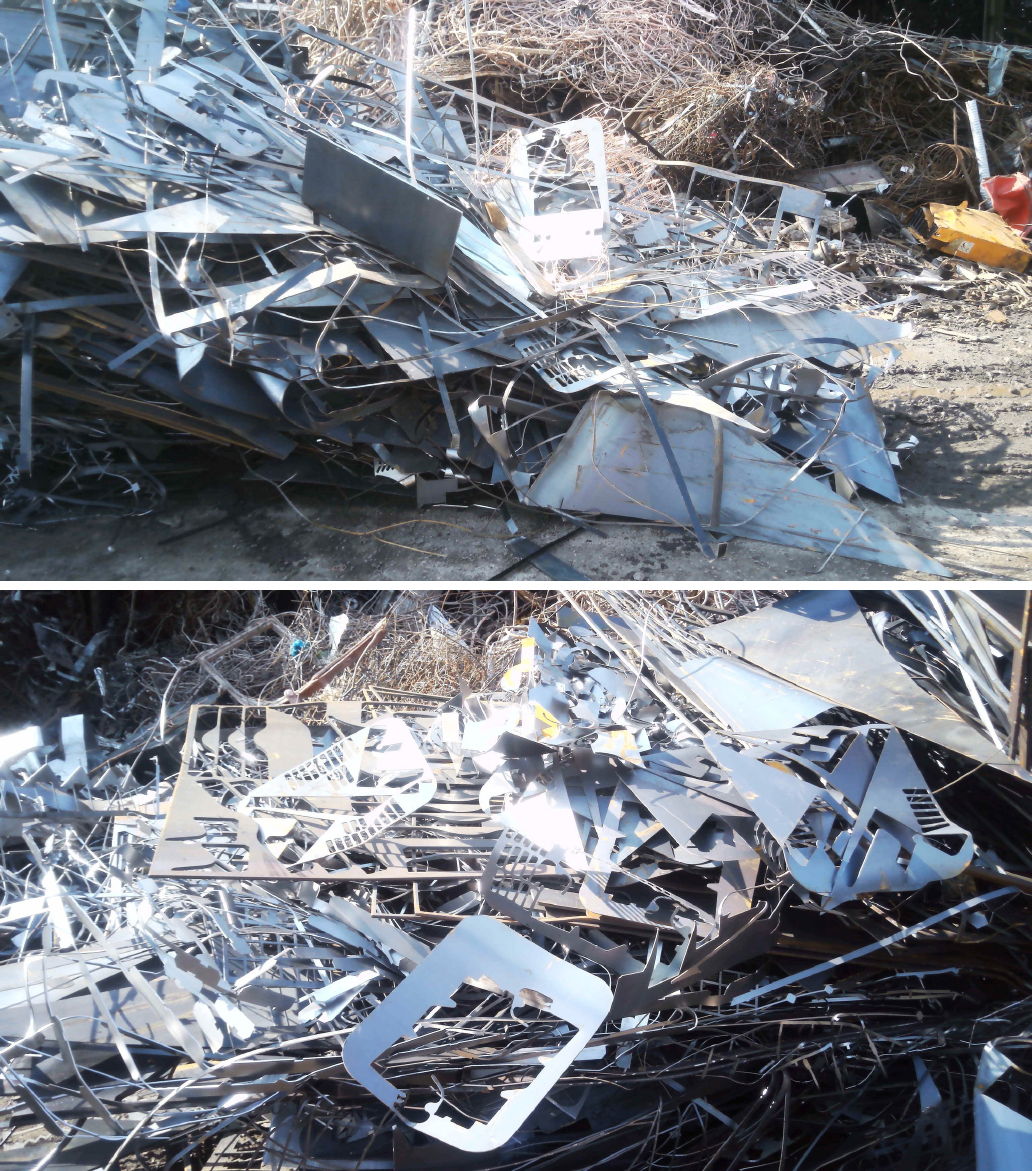}
\caption{Two of three retained views of one pile with a homogeneous
target: 100\% $>6$\,mm HMS. The seed-42 Mean pooling model assigns
99.97\% to Stamped steel using all three views (TV 99.98\%). The cut-out
plate shapes resemble stamping offcuts, but their thickness and the
operator's taxonomy decision cannot be established from appearance alone.}
\label{fig:supp-homogeneous-failures}
\end{figure}

Figure~\ref{fig:supp-homogeneous-failures} shows two views of the second
case. The cut-out plate shapes resemble stamping offcuts, but the operator
recorded entirely $>6$\,mm HMS. Plate thickness may explain this taxonomy
choice. Without measurements or independent reassessment, these views
cannot establish the cause of the disagreement.

%% file: supplementary_sections/revised_qualitative.tex
\clearpage
\section{Qualitative Evidence}
\label{sec:supp-centered-logits}

\begin{figure}[H]
\centering
\includegraphics[width=0.86\textwidth]{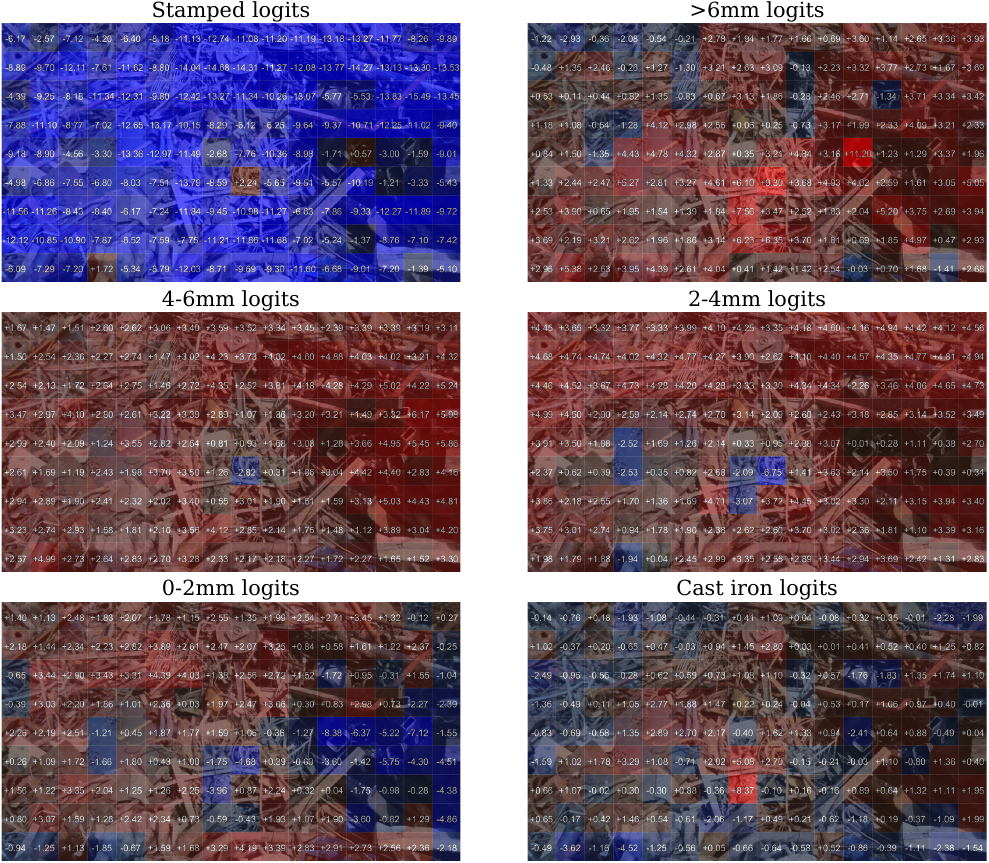}
\caption{Centered token contributions from the Mean pooling model
for one retained view. Red supports the named class relative to the average
class logit; blue opposes it. In panel order, the pile-level target is
$(0,70,15,10,5,0)\%$ and the prediction using all retained views is
$(0.00,37.48,32.15,24.48,3.27,2.63)\%$.}
\label{fig:supp-six-class-map}
\end{figure}

\paragraph{Centered token contributions.}
For a projected token $\mathbf z_t$ and class-head weights $\mathbf w_c$,
we display $(\mathbf w_c-\bar{\mathbf w})^\top\mathbf z_t$, where
$\bar{\mathbf w}=\frac16\sum_{c=1}^{6}\mathbf w_c$.
Subtracting the shared weight component leaves softmax probabilities
unchanged. Red and blue show positive and negative contributions relative
to the average class logit, and the spatially constant head bias is omitted.
One token may support several classes, for example where materials overlap.
These are logit contributions, not class-exclusive segmentation or local
mass measurements (Fig.~\ref{fig:supp-six-class-map}).

\begin{figure}[H]
\centering
\includegraphics[width=0.72\textwidth]{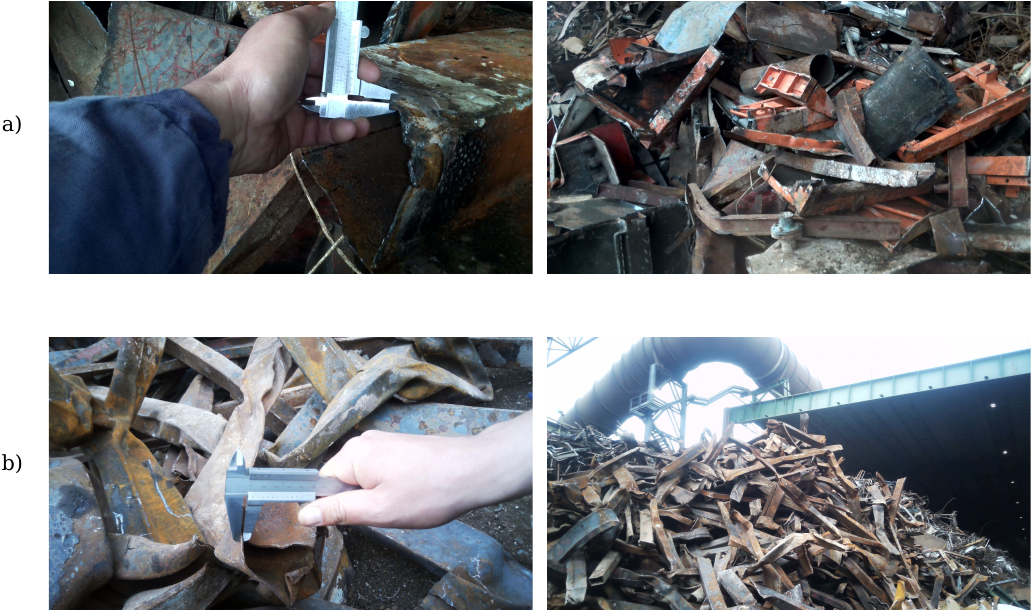}
\caption{Thickness measurements during grading. (a) Target/Mean/oracle
$>6$\,mm shares are 70/98.0/93.8\%; (b) target/Mean/oracle 2--4\,mm
shares are 70/15.5/20.4\%. Reweighting moves both predictions toward the
operator estimates but leaves substantial thickness-class disagreement.
The local measurements do not establish pile-level mass fractions.}
\label{fig:supp-ruler-cases}
\end{figure}

\section{Backbone Implementation}
\label{sec:supp-backbones}
All backbones in main-paper Table~3 are frozen. The DINOv2 models and CLIP
ViT-L/14 use $896\times504$ inputs, while DINOv3 ViT-L/16 and ConvNeXt-V2-L use
$1024\times576$. The ViTs produce $64\times36$ spatial grids, pooled in
$4\times4$ blocks to $16\times9$. CLIP uses its own RGB normalization and
interpolated positional embeddings. The ConvNeXt-V2-L checkpoint is FCMAE
pretrained and ImageNet fine-tuned. Its stride-32 $32\times18$ map is pooled
in $2\times2$ blocks to $16\times9$. It uses checkpoint-specific normalization
and direct bicubic resizing, without the checkpoint's native crop. Feature
dimensions differ across backbones, but all are projected to 256 dimensions.

\clearpage
\section{Pooled Risk--Coverage}
\label{sec:supp-pooled-risk}
\begin{figure}[H]
\centering
\includegraphics[width=0.6\textwidth]{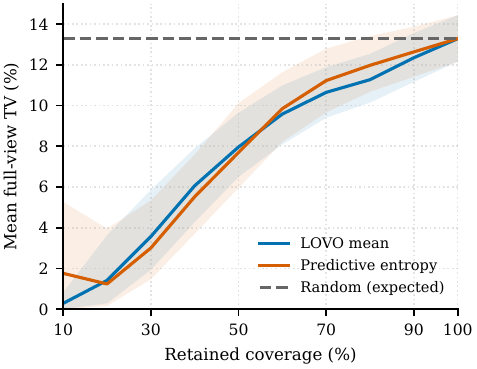}
\caption{Pooled risk--coverage for the seed-42 Mean pooling model on all 567
$F_1$--$F_4$ piles with at least two views, including homogeneous targets;
33 single-view piles are excluded. Shading shows pointwise 95\% pile-bootstrap
intervals. Both scores reduce average retained error, but the failure in
Fig.~\ref{fig:supp-homogeneous-failures} remains among entropy's most-confident 10\%.}
\label{fig:supp-risk-all}
\end{figure}

Figure~\ref{fig:supp-risk-all} extends the main-paper heterogeneous-pile
analysis to all eligible piles, including homogeneous targets.

